\documentclass[11pt]{article}
\usepackage{subcaption}

\usepackage{naaclhlt2016}
\usepackage{times}
\usepackage[normalem]{ulem}
\usepackage[utf8]{inputenc}
\usepackage{graphicx}
\usepackage{amsmath}
\usepackage{amssymb}
\usepackage{nicefrac}
\usepackage{xparse}
\usepackage{multirow}
\usepackage{booktabs}
\usepackage[scientific-notation=true]{siunitx}
\usepackage{url}
\usepackage{paralist}
\usepackage{comment}
\usepackage{tikz}
\usepackage{microtype}
\usepackage{gb4e}
\naaclfinalcopy
\makeatletter
\newcommand{\@BIBLABEL}{\@emptybiblabel}
\newcommand{\@emptybiblabel}[1]{}
\makeatother
\usepackage{hyperref}
\hypersetup{ colorlinks=false, pdfborder={0 0 0}, }

\DeclareMathOperator*{\argmax}{arg\,max}
\DeclareTextFontCommand{\emph}{\bfseries}
\usepackage{relsize}
\usepackage{xspace}

\newcommand{\m}{multilingual\xspace}

\newcommand{\x}{crosslingual\xspace}

\newcommand{\w}{word\xspace}

\newcommand{\rep}{representation\xspace}

\newcommand{\rs}{{\rep}s\xspace}
\newcommand{\wrs}{{\w} {\rep}s\xspace}

\newcommand{\mwrs}{{\m} {\w} {\rep}s\xspace}

\title{Bilingual Learning of Multi-sense Embeddings with Discrete Autoencoders}

\author{Simon Šuster \\
 University of Groningen\\
 Netherlands \\
 {\tt s.suster@rug.nl} \\ \And
 Ivan Titov\\
 University of Amsterdam \\
 Netherlands\\
 {\tt titov@uva.nl}\\
\And
 Gertjan van Noord\\
 University of Groningen\\
 Netherlands\\
 {\tt g.j.m.van.noord@rug.nl}
}
\begin{document}
\maketitle
\begin{abstract}
We present an approach to learning multi-sense word embeddings relying both on monolingual and bilingual information.
Our model consists of an encoder, which uses monolingual and bilingual context (i.e. a parallel sentence) to choose a sense for a given word, and a decoder which predicts
context words based on the chosen sense. The two components are estimated jointly.
We observe that the word representations induced from bilingual data outperform the monolingual counterparts across a range of evaluation tasks, even though crosslingual information is not available at test time.
\end{abstract}

\section{Introduction}
Approaches to learning word embeddings (i.e. real-valued vectors) relying on word context have received much attention in recent years, and the induced representations have been shown to capture
syntactic and semantic properties of words.
They have been evaluated intrinsically \cite{MikolovEtAl2013a,BaroniEtAl2014,LevyAndGoldberg2014b} and have also been used in concrete NLP applications to deal with word sparsity and improve generalization \cite{TurianEtAl2010,CollobertEtAl2011,BansalEtAl2014,PassosEtAl2014}.
While most work to date has focused on developing embedding models which represent a word with a single vector, some researchers have attempted to capture {\it polysemy} explicitly and have encoded properties of each word with multiple vectors \cite{HuangEtAl2012,TianEtAl2014,NeelakantanEtAl2014,ChenEtAl2014,LiAndJurafsky2015}.

In parallel to this work on multi-sense word embeddings, another line of research has
investigated integrating {\it \m} data, with largely two distinct goals in mind.
The first goal has been to obtain \rs for several languages in the same semantic space, which then enables the transfer of a  model (e.g., a syntactic parser) trained on annotated training data in one language to another language lacking this annotation \cite{KlementievEtAl2012,HermannAndBlunsom2014,GouwsEtAl2014,ChandarEtAl2014}.
Secondly, information from another language can also be leveraged to yield better first-language embeddings \cite{GuoEtAl2014}.
Our paper falls in the latter, much less explored category. 
We adhere to the view of \m learning as a means of language grounding \cite{FaruquiAndDyer2014,ZouEtAl2013,TitovAndKlementiev2012,SnyderAndBarzilay2010,NaseemEtAl2009}.
Intuitively, polysemy in one language can be at least partially resolved by looking at the translation of the word and its context in another language \cite{Kaji2003,NgEtAl2003,DiabAndResnik2002,Ide2000,DaganAndItai1994,BrownEtAl1991}.
Better sense assignment can then lead to better sense-specific word embeddings.

We propose a model that uses second-language embeddings as a supervisory signal in learning multi-sense \rs in the first language.
This supervision is easy to obtain for many language pairs as numerous parallel corpora exist nowadays.
Our model, which can be seen as an autoencoder with a discrete hidden layer encoding word senses, leverages bilingual data in its encoding part, while the decoder predicts the surrounding words relying on the predicted senses.
We strive to remain flexible as to the form of parallel data used in training and support both the use of word- and sentence-level alignments.

Our findings are:
\begin{itemize}
\item The second-language signal effectively improves the quality of multi-sense embeddings as seen on a variety of intrinsic tasks for English, with the results
superior to that of the baseline Skip-Gram model, even though the crosslingual information is not available at test time.
\item This finding is robust across several settings, such as varying dimensionality, vocabulary size and amount of data.
\item In the extrinsic POS-tagging task, the second-language signal also offers improvements over monolingually-trained multi-sense embeddings, however, the standard Skip-Gram embeddings turn out to be the most robust in this task.
\end{itemize}

We make the implementation of all the models as well as the evaluation scripts available at \url{http://github.com/rug-compling/bimu}.
\begin{figure}
\input{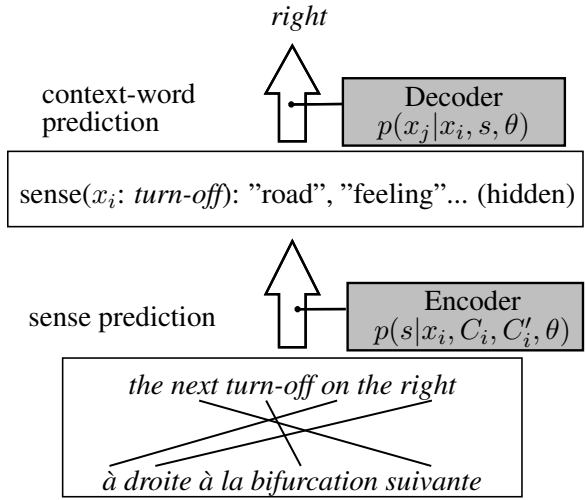}
\caption{Model schema: the sense encoder with bilingual signal and the context-word predictor are learned jointly.}
\end{figure}

\section{Word Embeddings with Discrete Autoencoders}
Our method borrows its general structure from neural autoencoders \cite{RumelhartEtAl1986,BengioEtAl2013}.
Autoencoders are trained to reproduce their input by first mapping
their input to a (lower dimensional) hidden layer and then predicting an approximation of
the input relying on this hidden layer.
In our case, the hidden layer is not a real-valued vector, but is a categorical variable encoding the sense of a word.
Discrete-state autoencoders have been successful in
several natural language processing applications, including POS tagging and word alignment~\cite{AmmarEtAl2014}, semantic role induction \cite{TitovAndKhoddam2015} and relation discovery \cite{MarcheggianiAndTitov2016}.

More formally, our model consists of two components: an {\it encoding} part which assigns a sense to a pivot word,
and  a {\it reconstruction} (decoding) part recovering context words based on the pivot word and its sense. As predictions are probabilistic (‘soft’), the reconstruction step involves summation over all potential word senses. %
The goal is to find embedding parameters which minimize the error in recovering context words based on the pivot word and the sense assignment. Parameters of both encoding and reconstruction are jointly optimized.
Intuitively, a good sense assignment should make the reconstruction step as easy as possible.
The encoder uses not only words in the first-language sentence to choose the sense but also, at training time, is conditioning its decisions on the words in the second-language sentence.
We hypothesize that the injection of crosslingual information will guide learning towards inducing more informative sense-specific word representations.
Consequently, using this information at training time would benefit the model even though crosslingual information is not available to the encoder at test time.

We specify the encoding part as a log-linear model:
\begin{align}\label{eq:encoder}
\nonumber
p(s|x_i,C_i,C'_i,\theta) & \propto \text{exp} \big( \varphi_{i,s}^\top  ( \frac{1-\lambda}{|C_i|} \sum_{j\in C_i} \gamma_j + \\ & \frac{\lambda}{|C'_i|} \sum_{k\in C'_i} \gamma'_k   )\big).
\end{align}
\noindent

To choose the sense $s\in\mathcal{S}$ for a word $x_i$, we use the bag of context words $C_i$ from the first language $l$, as well as the bag of context words $C'_i$ from the second language $l'$.\footnote{We have also considered a formulation which included a sense-specific bias $b_{x_i,s}\in\mathbb{R}$ to capture relative frequency of latent senses but it did not seem to affect performance.}
The context $C_i$ is defined as a multiset $C_i = \{x_{i-n},\ldots,x_{i-1},x_{i+1},\ldots,x_{i+n}\}$, including words around the pivot word in the window of size $n$ to each side.
We set $n$ to 5 in all our experiments.
The crosslingual context $C'_i$ is discussed in   \autoref{sec:align}, where we either rely on word alignments or use the entire second-language sentence as the context.
We distinguish between sense-specific embeddings, denoted by $\varphi\in\mathbb{R}^d$, and generic sense-agnostic ones, denoted $\{\gamma,\gamma'\} \in\mathbb{R}^d$ for first and second language, respectively.
The number of sense-specific embeddings is the same for all words.
We use $\theta$ to denote all these embedding parameters.
They are learned jointly, with the exception of the pre-trained second-language embeddings.

The hyperparameter $\lambda\in\mathbb{R}, 0\leq\lambda\leq 1$  weights  the contribution of each language.
Setting $\lambda=0$ would drop the second-language component and use only the first language.
Our formulation allows the addition of new languages easily, provided that the second-language embeddings live in the same semantic space.

The reconstruction part predicts a context word $x_j$ given the pivot $x_i$ and the current estimate of its $s$:
\begin{equation}
p(x_j|x_i,s,\theta) = \frac{\text{exp}(\varphi_{i,s}^\top \gamma_j)}{\sum_{k\in|\mathcal{V}|}\text{exp}(\varphi_{i,s}^\top \gamma_k)},
\end{equation}\label{eq:sgsoftmax}
\noindent
where $|\mathcal{V}|$ is the vocabulary size.
This is effectively a Skip-Gram model \cite{MikolovEtAl2013a} extended to rely on senses.

\subsection{Learning and regularization}

As sense assignments are not observed during training, the learning objective includes marginalization over word senses and thus can be written as:
\begin{equation}
\nonumber
\sum_i \! \sum_{\! j\in C_{x_i}}\! \log \sum_{s\in\mathcal{S}} \! p(x_j|x_i,s,\theta)p(s|x_i,C_i,C'_i,\theta),
\end{equation}
\noindent
in which index $i$ goes over all pivot words in the first language, $j$ over all context words to predict at each $i$, and $s$ marginalizes over all possible senses of the word $x_i$.
In practice, we avoid the costly computation of the normalization factor in the softmax computation of Eq.\ (\ref{eq:sgsoftmax}) and  use negative sampling \cite{MikolovEtAl2013b} instead of $\log p(x_j|x_i,s,\theta)$:
\begin{equation}
\log \sigma (\varphi_{i,s}^\top \gamma_j) + \sum_{x\in N} \log \sigma (-\varphi_{i,s}^\top \gamma_x),
\end{equation}
\noindent
where $\sigma$ is the sigmoid non-linearity function and $\gamma_x$ is a word embedding from the sample of negative (noisy) words $N$. %
Optimizing the autoencoding objective is broadly similar to the learning algorithm defined for multi-sense embedding induction in some of the previous work \cite{NeelakantanEtAl2014,LiAndJurafsky2015}.
Note though that this previous work has considered only monolingual context.

We use a minibatch training regime and seek to optimize the objective function $L(\mathcal{B},\theta)$ for each minibatch $\mathcal{B}$.
We found that optimizing this objective directly often resulted in inducing very flat posterior distributions. We therefore use a form of posterior regularization~\cite{GanchevEtAl2010} where we can encode our prior expectations that the posteriors should be sharp.
The regularized objective for a minibatch is defined as
\begin{equation}
L(\mathcal{B},\theta) + \lambda_H \sum_{i\in\mathcal{B}} H(q_i),
\end{equation}
\noindent
where $H$ is the entropy function and $q_i$ are the posterior distributions from the encoder ($p(s|x_i,C_i,C'_i,\theta)$).
This modified objective can also be motivated from a variational approximation perspective, see Marcheggiani and Titov~\shortcite{MarcheggianiAndTitov2016} for details.
By varying the parameter $\lambda_H\in\mathbb{R}$, it is easy to control the amount of entropy regularization.
For $\lambda_H>0$, the objective is optimized with flatter posteriors, while $\lambda_H<0$ infers more peaky posteriors.
When $\lambda_H\rightarrow-\infty$, the probability mass needs to be concentrated on a single sense, resulting in an algorithm similar to hard EM.
In practice, we found that using hard-update training\footnote{I.e.\ updating only that embedding $\varphi_{i,s^*}$ for which ${s^*=\argmax_{s} p(s|x_i,C_i,C'_i,\theta)}$.}, which is closely related to the $\lambda_H\rightarrow-\infty$ setting, led to best performance.


\subsection{Obtaining \wrs}
At test time, we construct the \wrs  by averaging all sense embeddings for a word $x_i$ and weighting them with the sense expectations \cite{LiAndJurafsky2015}\footnote{Although our training objective has sparsity-inducing properties, the posteriors at test time are not entirely peaked, which makes weighting beneficial.}:
\begin{equation}
\omega_{i} = \sum_{s\in\mathcal{S}} p(s|x_i, C_i) \varphi_{i,s}. \label{avgExp}
\end{equation}
\noindent
Unlike in training, the sense prediction step here does not use the crosslingual context $C'_i$ since it is not available in the evaluation tasks.
  In this work, instead of marginalizing out the unobservable crosslingual context, we simply ignore it in computation.

Sometimes, even the first-language context is missing, as is the situation in many word similarity tasks.
In that case, we just use the uniform average, $\nicefrac{1}{|\mathcal{S}|}\sum_{s\in\mathcal{S}} \varphi_{i,s}$.

\section{Word affiliation from alignments}\label{sec:align}
In defining the crosslingual signal we draw on a heuristic inspired by Devlin et al.~\shortcite{DevlinEtAl2014}.
The second-language context words are taken to be the multiset of words around and including the pivot affiliated to $x_i$:
\begin{equation}
C'_i = \{x'_{a_i-m}, ..., x'_{a_i}, ..., x'_{a_i+m}\},
\end{equation}
where $x'_{a_i}$ is the word affiliated to $x_i$ and the parameter $m$ regulates the context window size.
By choosing $m=0$, only the affiliated word is used as $l'$ context, and by choosing $m=\infty$, the $l'$ context is the entire sentence ($\approx$uniform alignment).
To obtain the index $a_i$, we use the following:
\begin{compactenum}[1)]
\item If $x_i$ aligns to exactly one second-language word, $a_i$ is the index of the word it aligns to.
\item If $x_i$ aligns to multiple words, $a_i$ is the index of the aligned word in the middle (and rounding down when necessary).
\item If $x_i$ is unaligned, $C'_i$ is empty, therefore no $l'$ context is used.
\end{compactenum}
We use the cdec aligner \cite{DyerEtAl2010} to word-align the parallel corpora.

\section{Parameters and Set-up}\label{sec:param}
\subsection{Learning parameters}
We use the AdaGrad optimizer \cite{DuchiEtAl2011} with initial learning rate set to 0.1.
We set the minibatch size to 1000, the number of negative samples to 1, the sampling factor to 0.001 and the window size parameter $m$ to 5.
All the embeddings are 50-dimensional (unless specified otherwise) and initialized by sampling from the uniform distribution between $[-0.05,0.05]$.
We include in the vocabulary all words occurring in the corpus at least 20 times.
We set the number of senses per word to 3 (see further discussion in \autoref{sec:nsenses} and \autoref{sec:related}).
All other parameters with their default values can be examined in the source code available online.

\subsection{Bilingual data}
In a large body of work on \mwrs, Europarl \cite{Koehn2005} is the preferred source of parallel data.
However, the domain of Europarl is rather constrained, whereas we would like to obtain \wrs of more general language, also to carry out an effective evaluation on semantic similarity datasets where domains are usually broader.
We therefore use the following parallel corpora: News Commentary \cite{NewsCommentary2013} (NC), Yandex-1M\footnote{\url{https://translate.yandex.ru/corpus}} (RU-EN), CzEng 1.0 \cite{Czeng2012} (CZ-EN) from which we exclude the EU legislation texts, and GigaFrEn \cite{GigaFrEn2009} (FR-EN).
The sizes of the corpora are reported in Table \ref{tab:corpora}.
The \wrs trained on the NC corpora are evaluated only intrinsically due to the small sizes.

\begin{table}[ht!]
\begin{tabular}{l l r r}
Corpus             & Language & Words & Sent. \\
\cmidrule(lr){1-4}
NC & Fr, Ru, Cz, De, Es & 3-4 M  & .1-.2 M \\
RU-EN & Ru & 24 M & 1 M \\
CZ-EN & Cz & 126 M & 10 M \\
FR-EN & Fr & 670 M & 23 M \\
\end{tabular}
\caption{Parallel corpora used in this paper. The word sizes reported are based on the English part of the corpus. Each language pair in NC has a different English part, hence the varying number of sentences per target language.}
\label{tab:corpora}
\end{table}

\section{Evaluation Tasks}
We evaluate the quality of our \wrs on a number of tasks, both intrinsic and extrinsic.
\subsection{Word similarity}\label{sec:similarity}
We are interested here in how well the semantic similarity ratings obtained from embedding comparisons correlate to human ratings.
For this purpose, we use a variety of similarity benchmarks for English and report the Spearman $\rho$ correlation scores between the human ratings and the cosine ratings obtained from our \wrs.
The \emph{SCWS} benchmark \cite{HuangEtAl2012} is probably the most suitable similarity dataset for evaluating multi-sense embeddings, since it allows us to perform the sense prediction step based on the sentential context provided for each word in the pair.

The other benchmarks we use provide the ratings for the word pairs without context.
WS-353 contains 353 human-rated word pairs \cite{FinkelsteinEtAl2001}, while Agirre et al.~\shortcite{AgirreEtAl2009} separate this benchmark for similarity (WS-SIM) and relatedness (WS-REL).
The RG-65 \cite{RubensteinAndGoodenough1965} and the MC-30 \cite{MillerAndCharles1991} benchmarks contain nouns only.
The MTurk-287 \cite{RadinskyEtAl2011} and MTurk-771 \cite{HalawiEtAl2012} include word pairs whose similarity was crowdsourced from AMT.
Similarly, MEN \cite{BruniEtAl2012} is an AMT-annotated dataset of 3000 word pairs.
The YP-130 \cite{YangAndPowers2006} and Verb-143 \cite{BakerEtAl2014} measure verb similarity.
Rare-Word \cite{LuongEtAl2013} contains 2034 rare-word pairs.
Finally, SimLex-999 \cite{HillEtAl2014c} is intended to measure pure similarity as opposed to relatedness.
For these benchmarks, we prepare the \wrs by taking a uniform average of all sense embeddings per word.
The evaluation is carried out using the tool described in Faruqui and Dyer\ \shortcite{FaruquiAndDyer2014b}.
Due to space constraints, we report the results by averaging over all benchmarks (\emph{Similarity}), and include the individual results in the online repository.

\subsection{Supersense similarity}
We also evaluate on a task measuring the similarity between the embeddings---in our case uniformly averaged in the case of multi-sense embeddings---and a matrix of supersense features extracted from the English SemCor, using the \emph{Qvec} tool \cite{TsvetkovEtAl2015}.
 We choose this method because it has been shown to output scores that correlate well with extrinsic tasks, e.g.\ text classification and sentiment analysis.
We believe that this, in combination with word similarity tasks from the previous section, can give a reliable picture of the generic quality of word embeddings studied in this work.

\subsection{POS tagging}
As our downstream evaluation task, we use the learned \wrs to initialize the embedding layer of a neural network tagging model.
We use the same convolutional architecture as Li and Jurafsky~\shortcite{LiAndJurafsky2015}: an input layer taking a concatenation of neighboring embeddings as input, three hidden layers with a rectified linear unit activation function and a softmax output layer.
We train for 10 epochs using one sentence as a batch.
Other hyperparameters can be examined in the source code.
The multi-sense word embeddings are inferred from the sentential context (weighted average), as for the evaluation on the SCWS dataset.
We use the standard splits of the Wall Street Journal portion of the Penn Treebank: 0--18 for training, 19--21 for development and 22--24 for testing.

\section{Results}
We compare three embeddings models, Skip-Gram (\textsc{Sg}), Multi-sense (\textsc{Mu}) and Bilingual Multi-sense (\textsc{BiMu}), using our own implementation for each of them.
The first two can be seen as simpler variants of the \textsc{BiMu} model: in \textsc{Sg} we omit the encoder entirely, and in \textsc{Mu} we omit the second-language ($l'$) part of the encoder in Eq.~(\ref{eq:encoder}).
We train the \textsc{Sg} and the \textsc{Mu} models on the English part of the parallel corpora.
Those parameters common to all methods are kept fixed during experiments.
The values $\lambda$ and $m$ for controlling the second-language signal in \textsc{BiMu} are set on the POS-tagging development set (cf.\ \autoref{bilingualparams}).

\begin{table}[ht!]
\hfill{}
{\footnotesize
\begin{tabular}{l l c c c l}
 Task & Corpus & \textsc{Sg} & \textsc{Mu} & \textsc{BiMu} & {\tiny\textsc{BiMu}-\textsc{Sg}}\\ 

\cmidrule(lr){1-6}
\multirow{8}{*}{\emph{\rotatebox[origin=c]{90}{SCWS}}} & RU-EN & 54.8 & 57.3 & \emph{59.5} & 4.7_{0.9}^{9.8} \\ 
                               & CZ-EN & 51.2 & 54.0 & \emph{55.3}  & 4.1_{-0.6}^{8.8} \\ 
                               & FR-EN & 58.8 & 60.4 & \emph{60.5}  & 1.7_{-2.6}^{5.9} \\ 
\cmidrule(lr){2-6}
                               & FR-EN (NC) & 47.2 & 52.4 & \emph{54.3}  & 7.1_{2.2}^{12.0} \\ 
                               & RU-EN (NC) & 47.3 & \emph{54.0} & \emph{54.0}  & 6.7_{0.6}^{12.8} \\ 
                               & CZ-EN (NC) & 47.7 & \emph{52.1} & 51.9  & 4.2_{-2.0}^{10.3} \\ 
                               & DE-EN (NC) & 48.5 & 52.9 & \emph{54.0}  & 5.5_{-0.6}^{11.6} \\ 
                               & ES-EN (NC) & 47.2 & 53.2 & \emph{54.5}  & 7.3_{1.1}^{13.3} \\ 
\cmidrule(lr){1-6}
\multirow{8}{*}{\emph{\rotatebox[origin=c]{90}{Similarity}}} & RU-EN & 37.8 & 41.2 & \emph{46.3}  & \\
                               & CZ-EN & 39.5 & 36.9 & \emph{41.9}  & \\
                               & FR-EN & \emph{46.3} & 42.0 & 43.5  & \\
\cmidrule(lr){2-5}
                               & FR-EN (NC) & 17.9 & 26.0 & \emph{27.6}  & \\
                               & RU-EN (NC) & 19.3 & 27.3 & \emph{28.4}  & \\
                               & CZ-EN (NC) & 15.8 & \emph{26.6} & 25.4  & \\
                               & DE-EN (NC) & 20.7 & 28.4 & \emph{30.8}  & \\
                               & ES-EN (NC) & 19.9 & 27.2 & \emph{31.2}  & \\
\cmidrule(lr){1-5}
\multirow{3}{*}{\emph{\rotatebox[origin=c]{90}{Qvec}}} & RU-EN & 55.8 & 56.0 & \emph{56.5}   & \\
                               & CZ-EN & \emph{56.6} & 56.5 & 55.9   & \\
                               & FR-EN & 57.5 & 57.1 & \emph{57.6}  & \\ 
\cmidrule(lr){1-5}
\multirow{3}{*}{\emph{\rotatebox[origin=c]{90}{POS}}} & RU-EN & \emph{93.5} & 93.2 & 93.3   & \\ 
                               & CZ-EN & \emph{94.0} & 93.7 & \emph{94.0}  &  \\ 
                               & FR-EN & \emph{94.1} & 93.8 & 94.0  & \\ 

\end{tabular}}
\hfill{}
\caption{Results, per-row best in bold. \textsc{Sg} and \textsc{Mu} are trained on the English part of the parallel corpora.
In {\tiny\textsc{BiMu}-\textsc{Sg}}, we report the difference between \textsc{BiMu} and \textsc{Sg}, together with the 95\% CI of that difference.
The \emph{Similarity} scores are averaged over 12 benchmarks described in \autoref{sec:similarity}.
For POS tagging, we report the accuracy.}%
\label{tab:results}
\end{table}

The results on the \emph{SCWS} benchmark (Table~\ref{tab:results}) show consistent improvements of the \textsc{BiMu} model over \textsc{Sg} and \textsc{Mu} across all parallel corpora, except on the small CZ-EN (NC) corpus.
We have also measured the 95\% confidence intervals of the difference between the correlation coefficients of \textsc{BiMu} and \textsc{Sg}, following the method described in Zou~\shortcite{Zou2007}.
According to these values, \textsc{BiMu} significantly outperforms \textsc{Sg} on RU-EN, and on French, Russian and Spanish NC corpora.\footnote{I.e.\ counting those results in which the CI of the difference does not include 0.}

Next, ignoring any language-specific factors, we would expect to observe a trend according to which the larger the corpus, the higher the correlation score.
However, this is not what we find.
Among the largest corpora, i.e.\ RU-EN, CZ-EN and FR-EN, the models trained on RU-EN perform surprisingly well, practically on par with the 23-times larger FR-EN corpus.
Similarly, the quality of the embeddings trained on CZ-EN is generally lower than when trained on the 10 times smaller RU-EN corpus.
 One explanation for this might be different text composition of the corpora, with RU-EN matching the domain of the evaluation task better than the larger two corpora.
 Also, FR-EN is known to be noisy, containing web-crawled sentences that are not parallel or not natural language \cite{DenkowskiEtAl2012}.
 Furthermore, language-dependent effects might be playing a role: for example, there are signs of Czech being the least helpful language among those studied.
 But while there is evidence for that in all intrinsic tasks, the situation in POS tagging does not confirm this speculation.

 We relate our models to previously reported SCWS scores from the literature using 300-dimensional models in Table \ref{tab:resultsrelated}.
Even though we train on a much smaller corpus than the previous works,\footnote{For example, Li and Jurafsky~\shortcite{LiAndJurafsky2015} use the concatenation of Gigaword and Wikipedia with more than 5B words.} the \textsc{BiMu} model achieves a very competitive correlation score.

\begin{table}
\hfill{}
{\footnotesize
\begin{tabular}{l c}
Model (300-dim.) & SCWS \\ 
\cmidrule(lr){1-2}
\textsc{Sg} & 65.0 \\
\textsc{Mu} & 66.7 \\
\textsc{BiMu} & 69.0 \\
Chen et al.~\shortcite{ChenEtAl2014} & 68.4 \\
Neelakantan et al.~\shortcite{NeelakantanEtAl2014} & 69.3 \\
Li and Jurafsky~\shortcite{LiAndJurafsky2015} & 69.7 \\
\end{tabular}}
\hfill{}
\caption{Comparison to other works (reprinted), for the vocabulary of top-6000 words. Our models are trained on RU-EN, a much smaller corpus than those used in previous work.}
\label{tab:resultsrelated}
\end{table}

\begin{figure*}
\begin{subfigure}{0.5\textwidth}
\includegraphics[width=\linewidth]{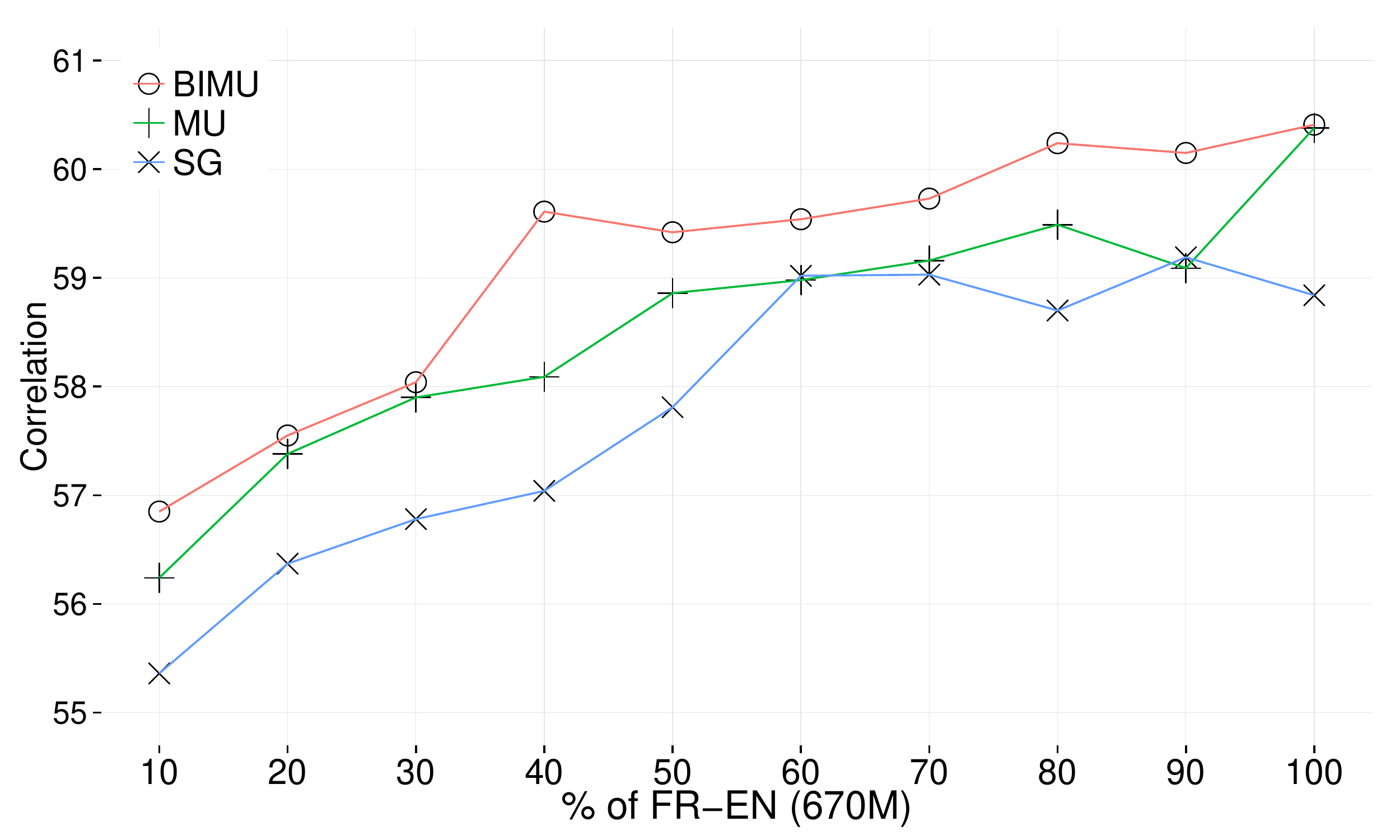}
\caption{} \label{fig:learncurve}
\end{subfigure}
\hspace*{\fill} 
\begin{subfigure}{0.5\textwidth}
\includegraphics[width=\linewidth]{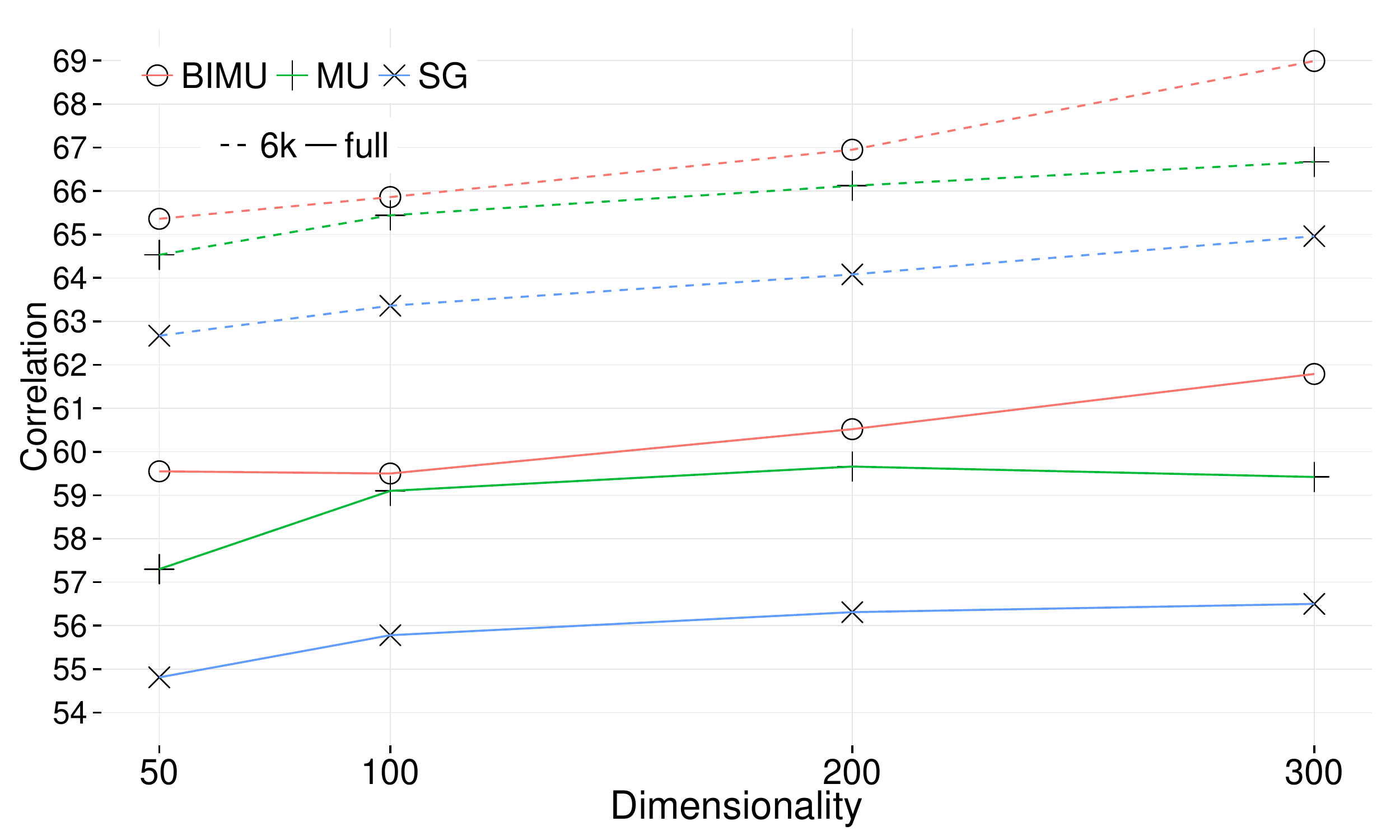}
\caption{} \label{fig:300d}
\end{subfigure}
\caption{(a) Effect of amount of data used in learning on the SCWS correlation scores. (b) Effect of embedding dimensionality on the models trained on RU-EN and evaluated on SCWS with either full vocabulary or the top-6000 words.} \label{fig:learncurve300d}
\end{figure*}

 The results on \emph{similarity} benchmarks and \emph{qvec} largely confirm those on SCWS, despite the lack of sentential context which would allow to weight the contribution of different senses more accurately for the multi-sense models.
 Why, then, does simply averaging the \textsc{Mu} and \textsc{BiMu} embeddings lead to better results than when using the \textsc{Sg} embeddings?
We hypothesize that the single-sense model tends to over-represent the dominant sense with its generic, one-vector-per-word representation, whereas
 the uniformly averaged embeddings yielded by the multi-sense models better encode the range of potential senses.
 Similar observations have been made in the context of selectional preference modeling of polysemous verbs~\cite{Greenberg2015}.

In \emph{POS} tagging, the relationship between \textsc{Mu} and \textsc{BiMu} models is similar as discussed above.
Overall, however, neither of the multi-sense models outperforms the \textsc{Sg} embeddings.
The neural network tagger may be able to implicitly perform disambiguation on top of single-sense \textsc{Sg} embeddings, similarly to what has been argued in Li and Jurafsky~\shortcite{LiAndJurafsky2015}.
The tagging accuracies obtained with \textsc{Mu} on CZ-EN and FR-EN are similar to the one obtained by Li and Jurafsky with their multi-sense model (93.8), while the accuracy of \textsc{Sg} is more competitive in our case (around 94.0 compared to 92.5), although they use a larger corpus for training the \wrs.

In all tasks, the addition of the bilingual component during training increases the accuracy of the encoder for most corpora, even though the bilingual information is not available during evaluation.

\subsection{The amount of (parallel) data}
Fig.\ \ref{fig:learncurve} displays how the semantic similarity as measured on SCWS evolves as a function of increasingly larger sub-samples from FR-EN, our largest parallel corpus.
The \textsc{BiMu} embeddings show relatively stable improvements over \textsc{Mu} and especially over \textsc{Sg} embeddings.
The same performance as that of \textsc{Sg} at 100\% is achieved by \textsc{Mu} and \textsc{BiMu} sooner, using only around 40/50\% of the corpus.

\subsection{The dimensionality and frequent words}
It is argued in Li and Jurafsky~\shortcite{LiAndJurafsky2015} that often just increasing the dimensionality of the \textsc{Sg} model suffices to obtain better results than that of their multi-sense model.
We look at the effect of dimensionality on semantic similarity in fig.\ \ref{fig:300d}, and see that simply increasing the dimensionality of the \textsc{Sg} model (to any of 100, 200 or 300 dimensions) is not sufficient to outperform the \textsc{Mu} or \textsc{BiMu} models.
When constraining the vocabulary to 6,000 most frequent words, the \rs obtain higher quality.
We can see that the models, especially \textsc{Sg}, benefit slightly more from the increased dimensionality when looking at these most frequent words.
This is according to expectations---frequent words need more representational capacity due to their complex semantic and syntactic behavior \cite{AtkinsAndRundell2008}.

\subsection{The role of bilingual signal}\label{bilingualparams}
\begin{figure*}
\begin{subfigure}{0.5\textwidth}
\includegraphics[width=\linewidth]{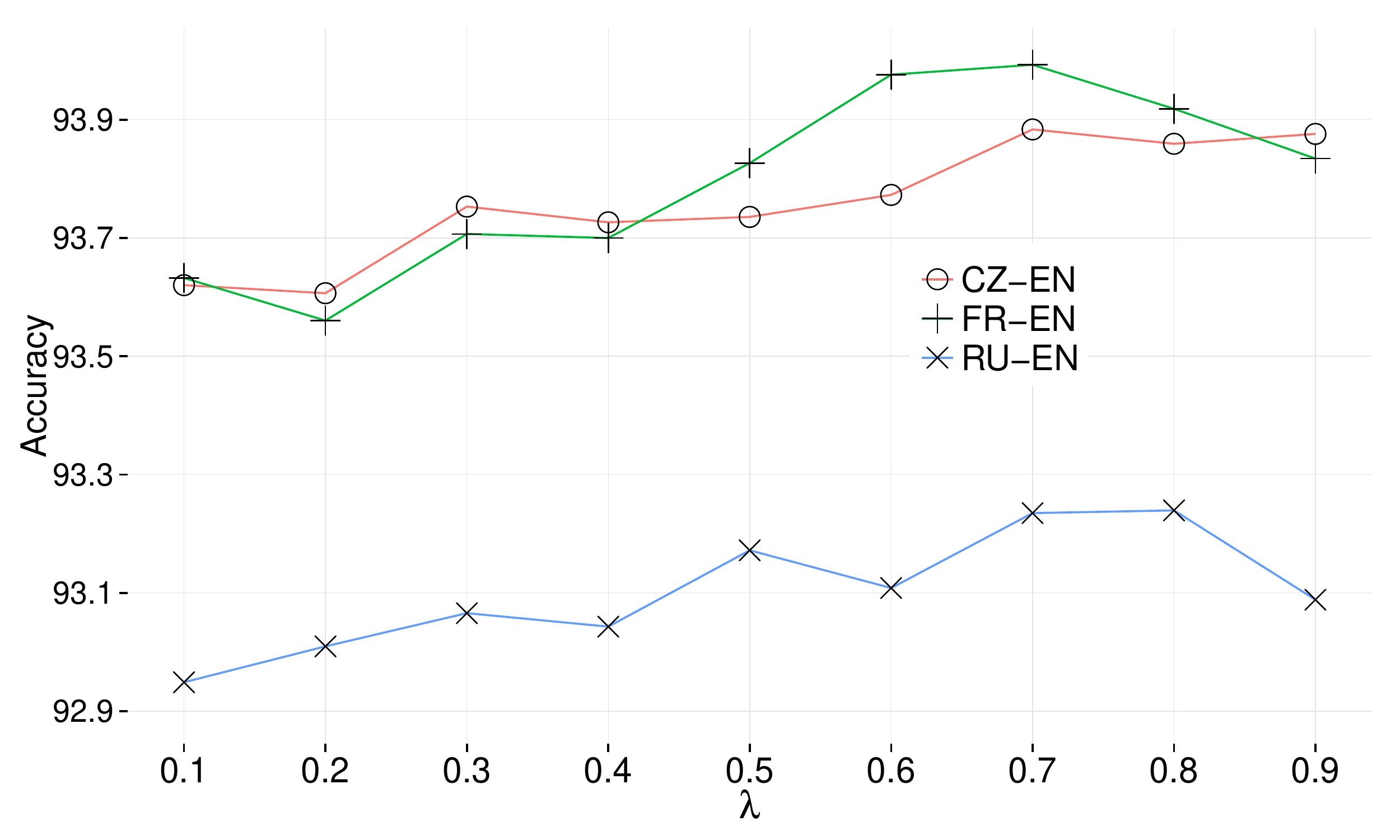}
\caption{} \label{fig:twoplots2}
\end{subfigure}
\hspace*{\fill} 
\begin{subfigure}{0.5\textwidth}
\includegraphics[width=\linewidth]{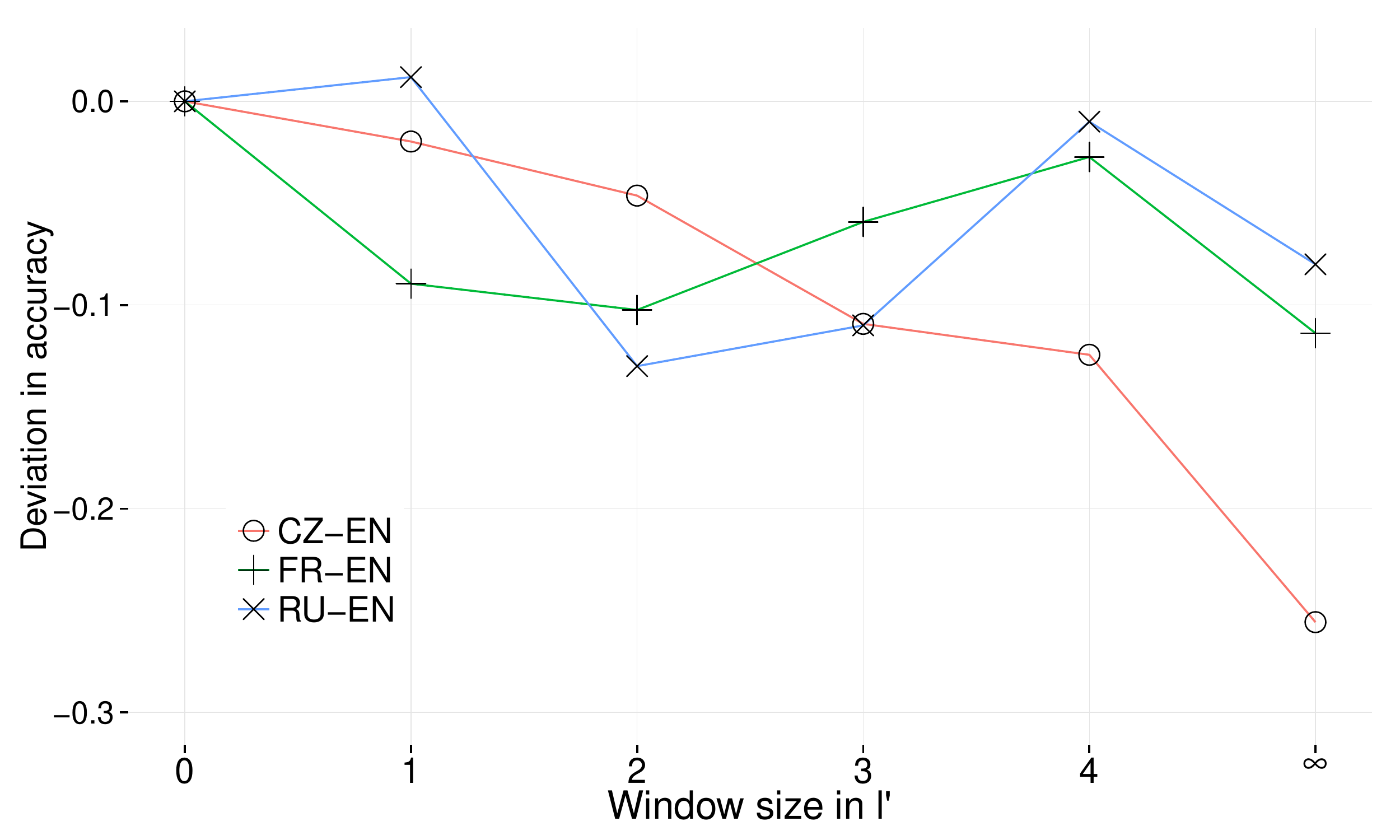}
\caption{} \label{fig:twoplots3}
\end{subfigure}
\caption{Controlling the bilingual signal. (a) Effect of varying the parameter $\lambda$ for controlling the importance of second-language context ($0.1$-least important, $0.9$-most important). (b) Effect of second-language window size $m$ on the accuracy. In both (a) and (b) the reported accuracies are measured on the POS-tagging development set. } \label{fig:twoplots}
\end{figure*}

The degree of contribution of the second language $l'$ during learning is affected by two parameters, $\lambda$ for the trade-off between the importance of first and second language in the sense prediction part (encoder) and the value of $m$ for the size of the window around the second-language word affiliated to the pivot.
Fig.\ \ref{fig:twoplots2} suggests that the context from the second language is useful in sense prediction, and that it should be weighted relatively heavily (around 0.7 and 0.8, depending on the language).

Regarding the role of the context-window size in sense disambiguation, the WSD literature has reported both smaller (more local) and larger (more topical) monolingual contexts to be useful, see e.g.\ Ide and V{\'e}ronis \shortcite{IdeAndVeronis1998} for an overview.
In fig.\ \ref{fig:twoplots3} we find that considering a very narrow context in the second language---the affiliated word only or a $m=1$ window around it---performs the best, and that there is little gain in using a broader window.
This is understandable since the $l'$ \rep participating in the sense selection is simply an average over all generic embeddings in the window, which means that the averaged \rep probably becomes noisy for large $m$, i.e.\ more irrelevant words are included in the window.
However, the negative effect on the accuracy is still relatively small, up to around $-0.1$ for the models using French and Russian as the second languages, and $-0.25$ for Czech when setting $m=\infty$.
The infinite window size setting, corresponding to the sentence-only alignment, performs well also on SCWS, improving on the monolingual multi-sense baseline on all corpora (Table \ref{tab:scwsinf}).

\begin{table}[h]
\hfill{}
{\footnotesize
\begin{tabular}{l r r r}
Model & RU-EN & CZ-EN & FR-EN \\
\cmidrule(lr){1-4}
\textsc{Mu} & 63.29 & 59.12 & 64.19 \\
\textsc{BiMu}, $m=\infty$ & \emph{65.61} & \emph{62.07} & \emph{64.36} \\ 
\end{tabular}}
\hfill{}
\caption{Comparison of SCWS correlation scores of \textsc{BiMu} trained with infinite $l'$  window to the \textsc{Mu} baseline (vocabulary of top-6000 words).}
\label{tab:scwsinf}
\end{table}

\subsection{The number of senses}\label{sec:nsenses}
In our work, the number of senses $k$ is a model parameter, which we keep fixed to 3 throughout the empirical study.
We comment here briefly on other choices of $k\in\{2,4,5\}$.
We have found $k=2$ to be a good choice on the RU-EN and FR-EN corpora (but not on CZ-EN), with an around $0.2$-point improvement over $k=3$ on SCWS and in POS tagging.
With the larger values of $k$, the performance tends to degrade.
For example, on RU-EN, the $k=5$ score on SCWS is about $0.6$ point below our default setting.

\section{Additional Related Work}\label{sec:related}
\noindent
{\bf Multi-sense models.} One line of research has dealt with sense induction as a separate, clustering problem that is followed by an embedding learning component \cite{HuangEtAl2012,ReisingerAndMooney2010}.
In another, the sense assignment and the embeddings are trained jointly \cite{NeelakantanEtAl2014,TianEtAl2014,LiAndJurafsky2015,BartunovEtAl2015}.
Neelakantan et al.~\shortcite{NeelakantanEtAl2014} propose an extension of Skip-Gram \cite{MikolovEtAl2013a} by introducing sense-specific parameters together with the $k$-means-inspired ‘centroid’ vectors that keep track of the contexts in
which word senses have occurred. They explore two model variants, one in which the number of senses is the same for all words, and another in which a threshold value determines the number of senses for each word.
 The results comparing the two variants are inconclusive, with the advantage of the dynamic variant being virtually nonexistent.
 In our work, we use the static approach.
Whenever there is evidence for less senses than the number of available sense vectors, this is unlikely to be a serious issue as the learning would concentrate on some of the senses, and these would then be the preferred predictions also at test time.
Li and Jurafsky~\shortcite{LiAndJurafsky2015} build upon the work of Neelakantan et al.\ with a more principled method for introducing new senses using the Chinese Restaurant Processes (CRP).
Our experiments confirm the findings of Neelakantan et al.\ that multi-sense embeddings improve Skip-gram embeddings on intrinsic tasks, as well as those of Li and Jurafsky, who find that multi-sense embeddings offer little benefit to the neural network learner on extrinsic tasks.
Our discrete-autoencoding method when viewed without the bilingual part in the encoder has a lot in common with their methods.

\noindent
{\bf Multilingual models.}
The research on using multilingual information in the learning of {\it multi-sense} embedding models is scarce.
Guo et al.~\shortcite{GuoEtAl2014} perform a sense induction step based on clustering translations prior to learning word embeddings.
Once the translations are clustered, they are mapped to a source corpus using WSD heuristics, after which a recurrent neural network is trained to obtain sense-specific \rs.
Unlike in our work, the sense induction and embedding learning components are entirely separated, without a possibility for one to influence another.
In a similar vein, Bansal et al.~\shortcite{BansalEtAl2012} use bilingual corpora to perform soft word clustering, extending the previous work on the monolingual case of Lin and Wu~\shortcite{LinAndWu2009}.
{\it Single-sense} \rs in the \m context have been studied more extensively \cite{LuEtAl2015,FaruquiAndDyer2014,HillEtAl2014,ZhangEtAl2014,FaruquiDyer2013,ZouEtAl2013}, with a goal of bringing the \rs in the same semantic space. A related line of work concerns the \x setting, where one tries to leverage training data in one language to build models for typically lower-resource languages \cite{HermannAndBlunsom2014,GouwsEtAl2014,ChandarEtAl2014,SoyerEtAl2014,KlementievEtAl2012,TackstromEtAl2012}.

The recent works of Kawakami and Dyer~\shortcite{KawakamiAndDyer2015} and Nalisnick and Ravi~\shortcite{NalisnickAndRavi2015} are also of interest.
The latter work on the infinite Skip-Gram model in which the embedding dimensionality is stochastic is relevant since it demonstrates that their embeddings exploit different dimensions to encode different word meanings.
Just like us, Kawakami and Dyer~\shortcite{KawakamiAndDyer2015} use bilingual supervision, but in a more complex LSTM network that is trained to predict word translations.
 Although they do not represent different word senses separately, their method produces \rs that depend on the context. 
In our work, the second-language signal is introduced only in the sense prediction component and is flexible---it can be defined in various ways and can be obtained from sentence-only alignments as a special case.

\section{Conclusion}
We have presented a method for learning multi-sense embeddings that performs sense estimation and context prediction jointly.
Both mono- and bilingual information is used in the sense prediction during training.
We have explored the model performance on a variety of tasks, showing that the bilingual signal improves the sense predictor, even though the crosslingual information is not available at test time.
In this way, we are able to obtain \wrs that are of better quality than the monolingually-trained multi-sense \rs, and that outperform the Skip-Gram embeddings on intrinsic tasks.
We have analyzed the model performance under several conditions, namely varying dimensionality, vocabulary size, amount of data, and size of the second-language context.
For the latter parameter, we find that bilingual information is useful even when using the entire sentence as context, suggesting that sentence-only alignment might be sufficient in certain situations.

\section*{Acknowledgments}
We would like to thank Jiwei Li for providing his tagger implementation, and Robert Grimm, Diego Marcheggiani and the anonymous reviewers for useful comments.
The computational work was carried out on Peregrine HPC cluster of the University of Groningen.
The second author was supported by NWO Vidi grant 016.153.327.
\bibliographystyle{naaclhlt2016}
\bibliography{0latexLiterature.bib}

\end{document}